\documentclass{article}

\usepackage{arxiv}
\usepackage[utf8]{inputenc} 
\usepackage[T1]{fontenc}    
\usepackage{hyperref}       
\usepackage{url}            
\usepackage{booktabs}       
\usepackage{amsfonts}       
\usepackage{nicefrac}       
\usepackage{microtype}      
\usepackage{lipsum}
\usepackage{algorithm}
\usepackage{algpseudocode}
\usepackage[english]{babel}
\usepackage{amsthm}
\usepackage{caption}
\usepackage{subcaption}
\usepackage{amsmath}
\usepackage{scalerel}
\usepackage{multicol}
\usepackage{easyReview}
\usepackage{amssymb}
\usepackage{tikz}
\usetikzlibrary{arrows.meta,chains,decorations.pathreplacing}
\usetikzlibrary{matrix,backgrounds,positioning}

\theoremstyle{definition}
\newtheorem{exmp}{Example}[section]
\theoremstyle{definition}
\newtheorem{definition}{Definition}[section]

\let\originallesssim\lesssim
\let\originalgtrsim\gtrsim
\DeclareRobustCommand{\lesssim}{%
  \mathrel{\mathpalette\lowersim\originallesssim}%
}
\DeclareRobustCommand{\gtrsim}{%
  \mathrel{\mathpalette\lowersim\originalgtrsim}%
}
\makeatletter
\newcommand{\lowersim}[2]{%
  \sbox\z@{$#1<$}%
  \raisebox{-\dimexpr\height-\ht\z@}{$\m@th#1#2$}%
}
\makeatother

\title{Understanding Recurrent Neural Architectures by Analyzing and Synthesizing Long Distance Dependencies in Benchmark Sequential Datasets}

\author{
  Abhijit Mahalunkar\\
  ADAPT Research Centre\\
  Technological University Dublin\\
  Dublin, Ireland\\
  \texttt{abhijit.mahalunkar@tudublin.ie} \\
  \And
  John D. Kelleher \\
  ADAPT Research Centre\\
  Technological University Dublin\\
  Dublin, Ireland\\
  \texttt{john.d.kelleher@tudublin.ie} \\
}

\begin{document}
\maketitle

\begin{abstract}
In order to build efficient deep recurrent neural architectures, it is essential to analyze the complexity of long distance dependencies (LDDs) of the dataset being modeled. In this paper, we present detailed analysis of the dependency decay curve exhibited by various datasets. The datasets sampled from a similar process (e.g. natural language, sequential MNIST, Strictly \emph{k}-Piecewise languages, etc) display variations in the properties of the dependency decay curve. Our analysis reveal the factors resulting in these variations; such as (i) number of unique symbols in a dataset, (ii) size of the dataset, (iii) number of interacting symbols within a given LDD, and (iv) the distance between the interacting symbols. We test these factors by generating synthesized datasets of the Strictly \emph{k}-Piecewise languages. Another advantage of these synthesized datasets is that they enable targeted testing of deep recurrent neural architectures in terms of their ability to model LDDs with different characteristics. We also demonstrate that analysing dependency decay curves can inform the selection of optimal hyper-parameters for SOTA deep recurrent neural architectures. This analysis can directly contribute to the development of more accurate and efficient sequential models.
\end{abstract}

\keywords{Long Distance Dependencies \and Sequential Models \and Recurrent Neural Networks}

\section{Introduction}
 Recurrent Neural Networks (RNN) laid the foundation of sequential data modeling \cite{Elman1990}. However, recurrent neural architectures trained using backpropagation through time (BPTT) suffer from exploding or vanishing gradients \cite{Hochreiter1991,Hochreiter01gradientflow,Yoshua1994}. This problem presents a specific challenge in modeling sequential datasets which exhibit long distance dependencies (LDDs). LDDs describe an interaction between two (or more) elements in a sequence that are separated by an arbitrary number of positions. LDDs are related to the rate of decay of statistical dependence of two points with increasing time interval or spatial distance between them. This dependence can be computed using information theoretic measure i.e. \emph{Mutual Information} \cite{Cover1991,Paninski2003,Bouma2009,Lin2017}. For example, in English there is a requirement for subjects and verbs to agree, compare: ``\emph{The \textbf{dog} in that house \textbf{is} aggressive}" with ``\emph{The \textbf{dogs} in that house \textbf{are} aggressive}". One of the early attempts at addressing this issue was by \cite{ElHihi1995} who proposed a hierarchical recurrent neural network which introduced several levels of state variables, working at different time scales. Various other architectures were developed based on these principles \cite{chang2017dilated,campos2018skip}. Another well-known approach to addressing this challenges is the Long Short-Term Memory (LSTM) introduced by \cite{hochreiter1997lstm}. The LSTM architecture models LDDs by enforcing constant error flow through \emph{constant error carousels} within special units. More recently, attention and memory augmented networks has shown to deliver good performance in modeling LDDs \cite{Merity2016,Graves2014,Salton2017}, and transformers use self-attention to model LDDs \cite{vaswani_2017,Dai2019}.

A fundamental task of modelling sequential data is Language Modeling. A language model accepts a sequence of symbols and predicts the next symbol in the sequence. The accuracy of a language model is dependent on the capacity of the model to capture the LDDs in the data on which it is evaluated because an inability to model LDDs in the input sequence will result in  erroneous predictions. In this paper, we reviewed the SOTA language models to check their performance on these datasets. The standard evaluation metric for language models is \emph{perplexity}. Perplexity is the measurement of how well a language model predicts the next symbol, and the lower the perplexity of a model the better the performance of the model. There are a number of benchmark datasets used to train and evaluate language models: Penn TreeBanks (PTB) \cite{MarcusPTB}, WikiText2, WikiText103 \cite{Merity2016} and Hutter-Text (Text8 and Enwik8).

Our review of the language model SOTA revealed that most research on developing language models fail to explicitly analyze the the LDDs within the datasets used to train and evaluate the models. Motivated by this, the paper makes a number of research contributions. First, we argue that a key step in modeling sequential data is to understand the properties of the LDDs within the data. Second, we present a method to compute and analyze the dependency decay curve for any sequential dataset, and demonstrate this method on a number of datasets that are frequently used to benchmark the SOTA sequential models. Third, based on the analysis of the dependency decay curves, we observe that LDDs are far more complex than previously assumed, and depend on at least four factors: (i) number of unique symbols in a dataset, (ii) size of the dataset, (iii) number of interacting symbols within an LDD, and (iv) distance between the interacting symbols. Fourth, we demonstrate how understanding dependency decay curve can inform better hyper-parameter selection for current SOTA recurrent neural architectures, and also aid in understanding them. We demonstrate this by using Strictly \emph{k}-Piecewise (SP\emph{k}) languages as a benchmark task for sequential models. The motivation for using the SP\emph{k} language modelling task, is that the standard sequential benchmark datasets provide little to no control over the factors which directly contribute to dependency decay curve. By contrast, we can generate benchmark datasets with varying degrees of LDD complexity by modifying the grammar of the SP\emph{k} language  \cite{Rogers2010,Fu2011,Avcu2017}. 

\section{Related Work}
Mutual information has previously been used to compute LDDs. Two literary texts, Moby Dick by H. Melville and Grimm's tales were used to analyze maximum length of LDDs present in English text \cite{Ebeling_1994}. Correlations were found between few hundred letters. More specifically, strong dependence was observed (large $\alpha 1$) upto 30 characters indicating strong grammar, beyond which point the curve exhibited a long tail indicating weak dependence. Dependency decay curves were analyzed of Enwik8 \cite{Lin2017}. It was observed that LDDs with power-law correlations tend to be more difficult to model. They argued that LSTMs are capable of modeling sequential datasets exhibiting LDDs with power law correlations such as natural languages far more effectively than markov models; due to power-law decay of hidden state of the LSTM network controlled by the forget gate. In another experiment, it was observed that DNA nucleotides exhibited long-range power law correlations \cite{Peng1992,Melnik2014}. 

Formal Language Theory, primarily developed to study the computational basis of human language is now being used extensively to analyze rule-governed systems \cite{Chomsky1956,CHOMSKY1959,Fitch2012}. Formal languages have previously been used to train RNNs and investigate their inner workings. The Reber grammar \cite{REBER1967} was used to train various 1\textsuperscript{st} order RNNs \cite{6796678,118646}. The Reber grammar was also used as a benchmark dataset for LSTM models \cite{hochreiter1997lstm}. Regular languages, studied by Tomita \cite{Tomita1982}, were used to train 2\textsuperscript{nd} order RNNs to learn grammatical structures of the strings \cite{Watrous1991,Giles1992}. Regular languages are the simplest grammars (type-3 grammars) within the Chomsky hierarchy which are driven by regular expressions. Strictly \emph{k}-Piecewise languages are natural and can express some of the kinds of LDDs found in natural languages \cite{Jager2012,Heinz2010}. This presents an opportunity of using SP\emph{k} grammar to generate benchmark datasets \cite{Avcu2017,Mahalunkar2018}. LSTM networks were trained to recognize valid strings generated using SP\emph{2}, SP\emph{4}, SP\emph{8} grammar \cite{Avcu2017}. LSTM could recognize valid strings generated using SP\emph{2} and SP\emph{4} grammar but struggled to recognize strings generated using SP\emph{8} grammar, exposing the performance bottleneck of LSTM networks. It was also observed that by increasing the maximum length of the generated strings of SP\emph{2} language (increasing the length of LDDs), the performance of LSTM degraded \cite{Mahalunkar2018}.

\section{Computing Dependency Decay Curves of Natural Datasets}
\subsection{Information Theoretic Association Measures}
Mutual information is an information-theoretic measure of association that computes the dependence between two or more events or distributions \cite{shannon1949,fano_1961,Cover1991}. Pointwise Mutual Information (PMI) computes the dependence between single events $x$ and $y$ belonging to the discrete random variables $X$ and $Y$ respectively by measuring their individual distributions and their joint distributions. PMI is $i$:
\begin{equation}
i(x;y) = \log \frac{p(x,y)}{p(x)p(y)}
\end{equation}

Average Mutual Information (MI) computes the dependence between two random variables $X$ and $Y$ with marginal distributions $p(x)$ and $p(y)$ and with the joint distribution $p(x,y)$. MI is the average or expectation over all possible events or PMI. MI is $I(X;Y)$:
\begin{equation}
I(X;Y) = \sum_{x,y} p(x,y) \log \frac{p(x,y)}{p(x)p(y)}
\end{equation}

MI can also be expressed using the entropy of $X$ and $Y$, i.e. $H(X)$, $H(Y)$, and their joint entropy, $H(X,Y)$, as:
\begin{equation}
I(X;Y) = H(X) + H(Y) - H(X,Y)
\label{eq:mut-inf-h}
\end{equation}
\begin{equation}
H(X) = \log N - 1/N \sum_{s=1}^{V} N_s \psi(N_s)
\label{eq:entropy-adj}
\end{equation}
Here, $N_s$ is the frequency of unique symbol $s$, $N {=} {\sum} N_s$, $V$ is the number of unique symbols or the vocabulary size, and $\psi(N_s)$ is the \emph{digamma function} of $N_s$. Shannon's entropy is known to be biased, generally underestimating the true entropy from finite samples; that is why, in order to compensate for insufficient sampling we use Eq.~\ref{eq:entropy-adj} from \cite{2003physics7138G} to calculate entropy.

If $I(X;Y){=}0$, then $X$ and $Y$ are independent then $p(x)p(y){=}p(x,y)$. However, if $X$ and $Y$ are fully dependent, then $p(x){=}p(y){=}p(x,y)$, which results in the maximum value of $I(X;Y)$. Similarly, if $i(x;y){=}0$, then events $x$ and $y$ are independent. If $i(x;y)$ is positive, it indicates that the two words occur together frequently. If $i(x;y)$ is negative, it indicates that the two words never occur together.

\subsection{Dependency Decay Curve}\label{sec:LDD_algo}
A dependency decay curve describes how the mutual information between symbols in a dataset decays as the distance between the symbols increases. To compute the dependency decay curve of a dataset we need to compute MI at every distance $d$ in the dataset, where $d$ is the spacing between pair of symbols. This is achieved by designing random variables $X$ and $Y$, where $X$ holds the subsequence of the original sequence with index range $[0,LEN{-}1{-}D]$, and $Y$ holds the subsequence with index range $[D,LEN{-}1]$ for all $D \in d$; where $LEN$ is the size of the dataset or the original sequence. The figure below illustrates how $X$ and $Y$ are defined over a sequence when $d=2$.

\begin{tikzpicture}[
       start chain = going right,
     node distance = 0pt,
MyStyle/.style={draw, minimum width=2em, minimum height=2em, 
                outer sep=0pt, on chain},
  ]
\node [MyStyle] (1) {$0$};
\node [MyStyle] (2) {$1$};
\node [MyStyle] (3) {$2$};
\node [MyStyle] (4) {$3$};
\node [MyStyle] (5) {$4$};
\node [MyStyle] (6) {$\cdots$};
\node [MyStyle] (7) {$LEN{-}3$};
\node [MyStyle] (8) {$LEN{-}2$};
\node [MyStyle] (9) {$LEN{-}1$};
\draw[decorate,decoration={brace, amplitude=10pt, raise=5pt}]
  (1.north west) to node[black,midway,above= 15pt] {$X$ elements} (7.north east);%
\draw[decorate,decoration={brace, amplitude=10pt, raise=5pt, mirror}]
  (3.south west) to node[black,midway,below= 15pt] {$Y$ elements} (9.south east);%
\end{tikzpicture}

Next we define a random variable $XY$ that contains a sequence of paired symbols one from $X$ and one from $Y$, where the symbols in a pair have the same index in $X$ and $Y$. The figure below illustrates the definition of these pairs, each column  defines one $XY$ pair.

\begin{tikzpicture}[
       start chain = going right,
     node distance = 0pt,
MyStyle/.style={draw, minimum width=2em, minimum height=2em, 
                outer sep=0pt, on chain},
  ]
\node [MyStyle] (1) {$0$};
\node [MyStyle] (2) {$1$};
\node [MyStyle] (3) {$2$};
\node [MyStyle] (4) {$3$};
\node [MyStyle] (5) {$4$};
\node [MyStyle] (6) {$\cdots$};
\node [MyStyle] (7) {$LEN{-}5$};
\node [MyStyle] (8) {$LEN{-}4$};
\node [MyStyle] (9) {$LEN{-}3$};
\draw[decorate,decoration={brace, amplitude=10pt, raise=5pt}]
  (1.north west) to node[black,midway,above= 15pt] {$X$ Random Variable} (9.north east);%
\end{tikzpicture}

\begin{tikzpicture}[
       start chain = going right,
     node distance = 0pt,
MyStyle/.style={draw, minimum width=2em, minimum height=2em, 
                outer sep=0pt, on chain},
  ]
\node [MyStyle] (1) {$2$};
\node [MyStyle] (2) {$3$};
\node [MyStyle] (3) {$4$};
\node [MyStyle] (4) {$5$};
\node [MyStyle] (5) {$6$};
\node [MyStyle] (6) {$\cdots$};
\node [MyStyle] (7) {$LEN{-}3$};
\node [MyStyle] (8) {$LEN{-}2$};
\node [MyStyle] (9) {$LEN{-}1$};
\draw[decorate,decoration={brace, amplitude=10pt, raise=5pt, mirror}]
  (1.south west) to node[black,midway,below= 15pt] {$Y$ Random Variable} (9.south east);%
\end{tikzpicture}

After this, we count the number of symbols that appear in $X$ and $Y$ (i.e., the size of the symbol vocabulary of $X$ and $Y$) these counts are stored in $K^X$, $K^Y$ respectively. Similarly, we count the number of unique pairs of symbols in $XY$ and store this in $K^{XY}$. We then obtain the frequency of each symbol in the vocabularies of $X$ and $Y$, giving us $N_i^X$ and $N_i^Y$; and the frequency of each of the pairs of symbols in $XY$, giving us $N_i^{XY}$. Using this information, and E.q.~\ref{eq:mut-inf-h} and \ref{eq:entropy-adj}, we calculate the mutual information $I(X;Y)$ at a distance $D$ in a sequence. We repeat this process for every $D {\in} d$, where $1{<}d{<}LEN$. The value of $I(X;Y)$ as a function of $d$ is the \emph{Dependency Decay Curve} $I(d)$. The area under the dependency decay curve describes the spread of dependencies along $d$. This curve is plotted on the log-log axis. The x-axis gives $d$, and the y-axis gives the MI at that $D$ (measured in \emph{nats}). Algorithm~\ref{ldd_algo} below explains the details. 

\begin{algorithm}
\begin{algorithmic}
 \For{$d\gets 1, |dataset|$}
  \State $X \gets dataset[0:|dataset|-D]$
  \State $Y \gets dataset[D:|dataset|]$
  \State $XY \gets$ zero-matrix of size ($K^X,K^Y$)
  \For{$i\gets 0,|X|$}
    \State Increment $XY[X[i],Y[i]]$
  \EndFor
  \State Compute $N_i^X$, $N^X$, $K^X$ for $X$
  \State Compute $N_i^Y$, $N^Y$, $K^Y$ for $Y$
  \State Compute $N_i^{XY}$, $N^{XY}$, $K^{XY}$ for $XY$
  \State Compute $H(X)$, $H(Y)$ and $H(X,Y)$ using Eq.~\ref{eq:entropy-adj}
  \State $I[D]\gets H(X)+H(Y)-H(X,Y)$
 \EndFor
 \State Return I[d]
\caption{Computing dependency decay curves}\label{ldd_algo}
\end{algorithmic}
\end{algorithm}

\subsection{Dependency Decay Curves of Word-Based Datasets}\label{ssec:ldds_word-based}
\begin{figure}
\centering
\includegraphics[scale=0.5]{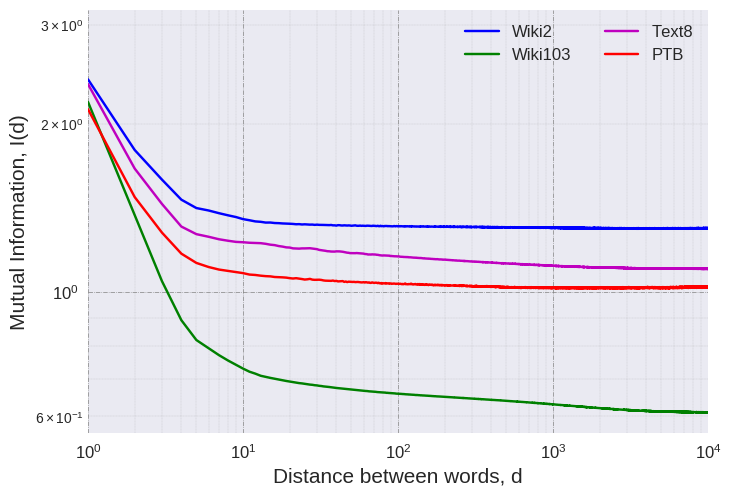}
\caption{Dependency decay: Word-based dataset}
\label{fig:lm_words}
\end{figure}

The dependency decay curves of word-based PTB, WikiText2, WikiText103 and Text8 datasets are in Fig.~\ref{fig:lm_words} and are plotted on a log-log axis where a straight line represents a power-law. Examining these curves, it is reasonable to describe each of these curves as combining multiple straight-line segments. The figure reveals that in each of these datasets the dependency decay follows multiple power-laws. Multiple power-laws that are combined together can be described using a \emph{broken power-law} relationship, an example of which is presented in Eq.~\ref{eqn:broken_p_l}. Here the broken power-law function $f(d)$ is comprised of three power-laws with exponents $\alpha_1$, $\alpha_2$, and $\alpha_3$ and corresponding constants $c_1$, $c_2$, and $c_3$, and combined at thresholds $d_{b1}$ and $d_{b2}$.
\begin{equation}\label{eqn:broken_p_l}
f(d) =
\left\{ 
  \begin{array}{ll}
    c_1 * d^{- \alpha_1} & d \leq d_{b1} \\
    c_2 * d^{- \alpha_2} & d_{b1} < d \leq d_{b2} \\
    c_3 * d^{- \alpha_3} & d > d_{b2}
  \end{array} 
\right.
\end{equation}

In general, the threshold or \emph{inflection point} that joins any two power-laws is of paramount importance as it indicates a change in a system where one phenomenon gives way to another. For example, in astronomy, the simplest afterglow light curves of the gamma-ray bursts can be described using a broken power-law. Here, the electron energy observed before and after the jet break (threshold) in the observer's frame exhibit different electron energy distribution index (i.e., different power laws) \cite{Rhoads_1999,Johannesson2006Afterglow}. For natural processes, this transition at the threshold is not sharp and is often replaced with a smoothly joined broken power-law. Eq.~\ref{eqn:broken_p_l} is of an ideal broken power-law with sharp discontinuities. To accommodate for smoothing, Eq.~\ref{eqn:broken_p_l} needs to be adjusted as below:
\begin{equation}\label{eqn:smooth_broken_p_l}
f_{sm}(d) =
\left\{ 
  \begin{array}{ll}
    c_1 * d^{- \alpha_1} & d \leq d_{b1} \\
    c_2 * d^{- \alpha_2} & d_{r1} < d \leq d_{b2} \\
    c_3 * d^{- \alpha_3} & d > d_{r2}
  \end{array} 
\right.
\end{equation}
Here, the first power-law i.e $c_1{*}d^{-\alpha_1}$ ends at $d_{b1}$ and second power-law i.e. $c_2{*}d^{-\alpha_2}$ begins at $d_{r1}$. The region between $d_{b1}{<}d{<}d_{r1}$ is a smooth curve joining the two power-laws where the slope of the curve in that region changes from $\alpha_1$ to $\alpha_2$. Hence this is the transition region. This also happens between the second power-law and third power-law i.e. $c_3{*}d^{-\alpha_3}$. Here, the second power-law ends at $d_{b2}$ and third power-law begins at $d_{r2}$ and the transition region is between $d_{b2}{<}d{<}d_{r2}$. Bringing the discussion back to the word-based English datasets, if we model the dependency decay exhibited by word-based dataset using equation~\ref{eqn:smooth_broken_p_l}, then $d$ describes the separation or distances between the symbols and $c_1$ describes the mutual information at a distance of one ($d{=}1$) (i.e. it describes the extent of the dependence of a word on its immediate neighbour). Furthermore, $\alpha_1$ describes the rate of decay of dependence between words as the separation between them $d$ increases, this rate of decay holds until the first inflection point ($d_{b1}$). Similarly, $c_2$ describes the mutual information at $d_{r1}$ and $\alpha_2$ describes the rate of decay of dependence between words as the separation between them increases for $d_{r1}{<}d{\leq}d_{b2}$, and so on. The presence of broken power-law signifies that the dependencies decay at different rates and these changes in decay rates may be caused by interactions between a number of different underlying phenomena.

\subsubsection{Fitting Word-Based Dependency Decay Curves}\label{ssec:fit}
To study the properties of LDDs in a given dataset, it is essential to obtain the properties of the dependency decay curves. Fitting to a dependency decay curve involves estimating the values for the thresholds (inflection points) between the component power-laws and the parameters of each component power-law. Unfortunately, we do not have access to a function that can generate a curve with three smooth broken power-laws. Consequently, the process we use involves first fitting a broken power-law combining two power-laws to an initial portion of the observed curve, and then fitting a second broken power-law combining two power-laws to the remainder of the curve. To implement this two-stage curve fitting process we use the \emph{SmoothlyBrokenPowerLaw1D} function (\emph{Astropy} library) to generate a smoothly broken power-law curve from a set of estimated parameters. 
\begin{equation}\label{eqn:smooth_broken}
    g(d) = A\left(\frac{d}{d_b}\right)^{-\alpha}\left[\frac{1}{2}+\frac{1}{2}\left(\frac{d}{d_b}\right)^{\frac{1}{\Delta}}\right]^{(\alpha-\alpha^{s})\Delta}
\end{equation}
where, $d_b$ is the threshold (inflection point) connecting the two power-laws, $\alpha$ is the index of the first power-law, $\alpha^{s}$ is the index of the second power-law, $A$ is the amplitude at the inflection, and $\Delta$ is a smoothness parameter which describes the smoothness of the transition between the two power-laws. In this context, the beginning of the transition region is denote by $d_l$ which marks where the transition phenomenon begins, $d_b$ marks the boundary where the relative dominance of the two phenomena switches, and the end of the transition region is denote by $d_r$, where the transition between the two phenomena is complete. Using these definitions for $d_l$, $d_r$, and $d_b$ the relationship between the smoothness parameter $\Delta$ and the transition between the two power-laws is:
\begin{equation}\label{eqn:smoothning}
    \log_{10} \frac{d_r}{d_b} = \log_{10} \frac{d_b}{d_l} \sim \Delta
\end{equation}
At values $d{\lesssim}d_l$ and $d{\gtrsim}d_r$ the model is approximately a simple power law with index $\alpha$ and $\alpha^s$ respectively. The two power laws are smoothly joined at values $d_l{<}d{<}d_r$, hence the $\Delta$ parameter sets the “smoothness” of the slope change. However, our experiments reveal that $d_l {\approx} d_b$, so for all practical purposes, $d_l$ can be replaced by $d_b$. Eq.~\ref{eqn:smooth_broken} does not take $c$ as a parameter and is calculated by the function \emph{SmoothlyBrokenPowerLaw1D}. 

\begin{figure}
\centering
\includegraphics[scale=0.4]{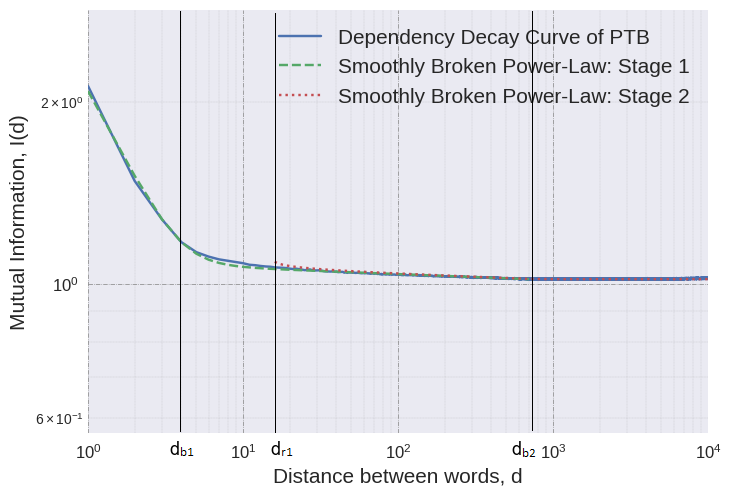}
\caption{Broken power-law fit: PTB dataset}
\label{fig:fit}
\end{figure}

In fitting the first broken power-law to a dependency decay curve we selected $I(d)$ where $1{<}d{<}700$. This selection was done by examining Fig.~\ref{fig:lm_words} and selecting the region of the curve that we were confident included one inflection point and two power-laws. We then generated estimates for the values of $\alpha_1$, $\alpha_1^s$, $d_{b1}$, $A_1$, and $\Delta_1$, and these estimates were passed as parameters to the Eq.~\ref{eqn:smooth_broken} to generate a smoothly broken power-law curve and obtain $c_1$, $d_{l1}$, and $d_{r1}$ values. To fit the second broken power-law we used the same process as we used to fit the first broken power-law with the distinction that we fitted the second broken power-law to the region of the dependency decay curve $I(d)$ where $d_{r1}{<}d{<}10000$. Here we obtain the value of the parameters of $c_2$, $\alpha_2$, $\alpha_2^s$, $A_2$, $\Delta_2$, $d_{l2}$, $d_{r2}$, and $d_{b2}$. Finally, the generated curve and the selected portion of the relevant dependency decay curve $I(d)$ were checked for the goodness of fit using the \emph{2-Sample Kolomogrov-Smirnov Test} (\emph{p-value}${<}0.05$). The fitting process is displayed in Fig.~\ref{fig:fit}. The power-law parameters and inflection points are listed in table~\ref{table:lm_words}. Note that we do not retain the values for $d_{l2}$ and $d_{r2}$ from the second broken power-law because we found that for all the datasets the inflection point between the two power-laws in the region was very sharp i.e., $d_{b2} {\approx} d_{l2} {\approx} d_{r2}$.

To illustrate how closely the two fitted broken power-laws match the dependency decay curves, Fig.~\ref{fig:fit} illustrates the fit between the dependency decay curve of PTB and the two smoothly broken-power law curves generated using the above process. In this figure the positions of $d_{b1}$, $d_{r1}$ (end of the transition region between the first and second power-law), and $d_{b2}$ are shown. Notice that there is a small discontinuity at $d_{r1}$ which marks the joint between the two phases of curve-fitting. However, apart from this discontinuity it is apparent that the fitted curves closely match the curves.

\begin{table*}
\small
\centering
\begin{tabular}[t]{lccccccccccccc}
\toprule
Datasets & Vocabulary & Size & $c_1$ & $\alpha_1$ & $d_{b1}$ & $\Delta_1$ & $d_{r1}$ & $c_2$ & $\alpha_2$ & $d_{b2}$ \\
\midrule
PTB & $10000$ & $1085779$ & $2.12$ & $0.4786$ & $4$ & $0.275$ & $15$ & $1.1716$ & $0.01$ & $728$ \\
Text8 & $253854$ & $17005208$ & $2.35$ & $0.4811$ & $4$ & $0.228$ & $12$ & $1.309$ & $0.016$ & $951$ \\
Wiki2 & $33278$ & $2551843$ & $2.38$ & $0.421$ & $4$ & $0.346$ & $23$ & $1.463$ & $0.0028$ & $2203$ \\
Wiki103 & $267735$ & $103690236$ & $2.19$ & $0.6931$ & $5$ & $0.35$ & $20$ & $0.821$ & $0.019$ & $2660$ \\
\bottomrule
\end{tabular}
\caption{Broken power-law fitting parameters for word-based datasets}\label{table:lm_words}
\end{table*}

\subsubsection{Analysis of Word-Based Datasets}
Here, we will discuss how the characteristics of LDDs, observed using the dependency decay curves result in perplexity values of different recurrent neural architectures. Table~\ref{table:lm_sota} lists the perplexity scores for test and valid sets for PTB, WikiText2 and WikiText103. There is a general trend in that model evaluations on WikiText103 tend to result in lower perplexity scores followed by WikiText2 and then PTB across a model. We are interested in understanding how the characteristics of the dataset (in particular properties of the LDDs in the data) affect the performance of recurrent neural architectures. With this goal in mind we analysed the attributes of the word-based natural language datasets. In order to examine the characteristics of the LDDs within each of these datasets we plotted the dependency decay curves indicating the decay of mutual information within the dataset as the spacing between the words is increased.

Dependency decay curves of the word-based natural language datasets follow expected trends \cite{Ebeling_1994}. It is seen that dependencies decay with a broken power-law \cite{Lin2017} as explained in section~\ref{ssec:fit}. For word-based datasets, we observe that $\alpha 1 {>} \alpha 2$. The higher value of $\alpha 1$ is due to a higher rate of reduction in the frequency of contextually correlated words in a sequence, as the spacing between them increases. This signifies the presence of a strong grammar. This strong dependence is observed between words at a distance up to $4$ across various datasets. For a given dataset, beyond the point of inflection, it is understood that the pairs are not contextually correlated which results in a flatter curve or lower value of $\alpha 2$. This analysis enables us to approximate the \emph{contextual boundary} of the natural language data. Also, the absolute of value of mutual information is an indicator of the degree of the short and long distance dependencies present in a dataset. The fact that our above analysis of the English datasets found a very large value for $\alpha 1$ indicates that a dataset with good distribution of English text will exhibit a high value of mutual information at lower values of $d$ followed by a steep decay of mutual information. As discussed above, we noted a trend in the results reported across the standard benchmark datasets where WikiText103 tended to deliver the best perplexity score followed by WikiText2 and PTB. Our analysis of the dependency decay curves provide an explanation for this trend. Language models have very good performance on WikiText103 due to the fact that they can take advantage of large $\alpha 1$ and very low mutual information in the flat region. Furthermore, language models marginally outperform on WikiText2 as compared to the PTB due to the higher mutual information at lower values of $d$.

\begin{table*}
\centering
\begin{tabular}{lccc}
\toprule
Model & PTB & WikiText2 & WikiText103 \\
\midrule
FRAGE + AWD-LSTM-MoS + dynamic eval \cite{Gong2018} & $47.38, 46.54$ & $40.85, 39.14$ & - \\
AWD-LSTM-DOC x5 \cite{Takase2018} & $48.63, 47.17$ & $54.19, 53.09$ & - \\
AWD-LSTM-MoS + dynamic eval \cite{Yang2017}* & $48.33, 47.69$ & $42.41, 40.68$ & - \\
AWD-LSTM + dynamic eval \cite{Krause2018}* & $51.6, 51.1$ & $46.4, 44.3$ & - \\
AWD-LSTM + continuous cache pointer \cite{Merity2017}* & $53.9, 52.8$ & - & - \\
AWD-LSTM-DOC \cite{Takase2018} & $54.12, 52.38$ & $53.8, 52.0$ & - \\
AWD-LSTM-MoS + finetune \cite{Yang2017} & $56.54, 54.44$ & - & - \\
AWD-LSTM-MoS \cite{Yang2017} & $58.08, 55.97$ & $63.88, 61.45$ & - \\
AWD-LSTM \cite{Merity2017}, 2017) & $60.0, 57.3$ & $68.6, 65.8$ & - \\
Transformer with tied adaptive embeddings \cite{Baevski2018adaptive} & - & - & $19.8, 20.5$ \\
LSTM + Hebbian + Cache + MbPA \cite{Jack2018} & - & - & $29.0, 29.2$\\
\bottomrule
\end{tabular}
\caption{Perplexity scores (test and valid) of SOTA word-based language models}\label{table:lm_sota}
\end{table*}

\subsection{Dependency Decay Curves of Character-Based Datasets}\label{ssec:ldds_char-based}
\begin{figure}
\centering
\includegraphics[scale=0.5]{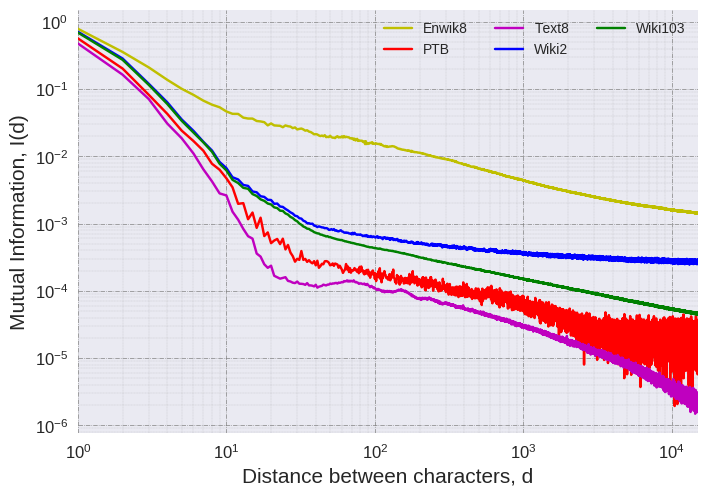}
\caption{Dependency decay: Character-based dataset}
\label{fig:lm_chars}
\end{figure}

In this section, we will study the dependency decay curve of character-based PTB, WikiText2, WikiText103, Text8, and Enwik8 datasets. They are displayed in Fig.~\ref{fig:lm_chars}. Dependency decay curves of character-based natural language datasets follow expected trends \cite{Montemurro2002}. As observed in word-based datasets, even the character-based datasets exhibit broken power-law dependence decay, an exception being Enwik8 (which we will discuss later). For character-based datasets, strong dependence (steeper dependence decay) is observed between characters at a distance up to $30$; beyond which it follows a long tail indicating lower dependence. Enwik8 displays a different decay characteristics than the other datasets as this dataset is made up of XML format and not English text. When we fit the dependency decay curve, we observe that it is made up of 4 constituent power-laws.

\subsection{Dependency Decay Curves of Mobility Dataset}\label{fig:ldds_mobility}
\begin{figure}
\centering
\includegraphics[scale=0.5]{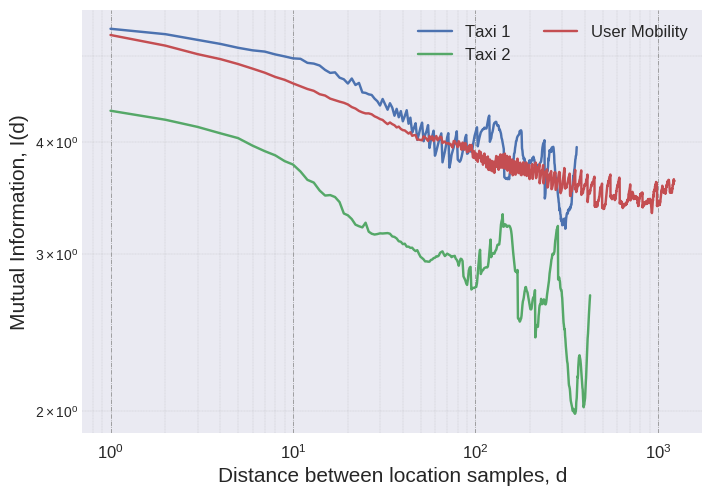}
\caption{Dependency decay: Human mobility dataset}
\label{fig:mobility}
\end{figure}

We also computed the dependency decay curves of the GPS trajectory dataset collected in Geolife project (Microsoft Research Asia) by 178 users in a period of over four years (from April 2007 to October 2011) and was plotted in Fig.~\ref{fig:mobility}. A GPS trajectory of this dataset is represented by a sequence of time-stamped points, each of which contains the information of latitude, longitude and altitude which was converted to a unique location number. These trajectories were recorded by different GPS loggers and GPS phone \cite{Geolife2011}. Upon analyzing the plot of the dependency decay curves in this data, it's evident that human mobility also has power law-decay suggesting the presence of LDDs.

\subsection{Dependency Decay Curves of Sequential MNIST Dataset}\label{ssec:ldds_sequential_mnist}
\begin{figure}
\centering
\includegraphics[scale=0.5]{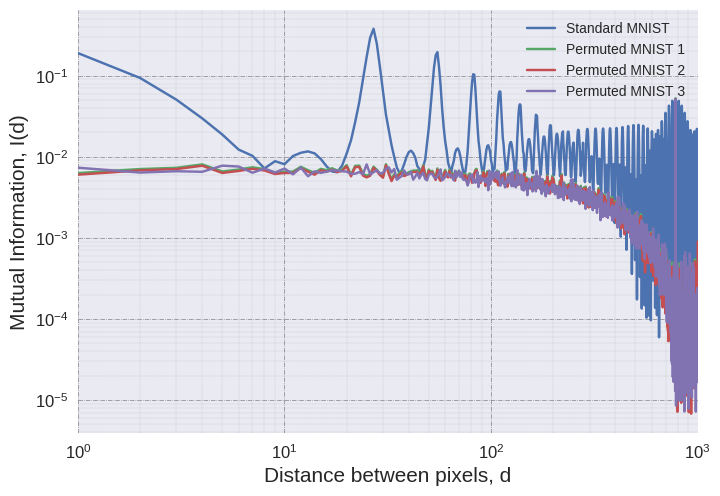}
\caption{Dependency decay: Unpermuted \emph{sequential MNIST} \& permuted \emph{sequential MNIST} using multiple seeds}
\label{fig:mnist}
\end{figure}

Sequential MNIST is widely used to evaluate recurrent neural architectures. It contains $240000$ training and $40000$ test images. Each of these is $28$x$28$ pixels in size, and each pixel can take one of $256$ pixel values. In order to use them in a sequential task, the $2$D images are converted into a $1$D vector of $784$ pixels by concatenating all the rows of the pixels. This transformation results in pixel dependencies which span up to approximately $28$ pixels. These dependencies arise due to high correlation of a pixel with its neighboring pixels. The structure of the Sequential MNIST dataset is such that its dependency decay curve is likely to contain regular peaks and troughs. We plot the dependency decay curve of the unpermuted and permuted sequential MNIST datasets in Fig.~\ref{fig:mnist}. 

Standard sequential MNIST exhibits high MI at $d{=}1$ indicating strong dependencies at close proximity. The dependencies then decay as a function of power-law. Hence, in-order to fully capture these dependencies, the recurrent neural architectures should maintain gradients/attention across multiple timescales as a function of power-law to accurately model these dependencies. However, we see peaks of mutual information at intervals of $28$ due to pixel dependencies. The regular peaks in the decay curve indicate that the span of the dependencies lie within $d{\approx}28$. We generated permuted versions of the sequential MNIST dataset with multiple seeds for use as a comparator with the unpermuted sequential MNIST. When we examine the dependency decay curves of permuted MNIST datasets, we observe that the dependencies are substantially less between close-by symbols (pixels in this case), e.g. for $d{=}1$ the green, red and purple lines are much lower than the blue line. This is a result of permutations applied to the data which disrupt spatial dependencies. Another impact of this disruption is the relatively flat curve for $d{<}300$ which indicates an absence of spatial dependencies. In-order to model these datasets that exhibit a relatively flat curve, the recurrent neural architectures requires uniform distribution of attention/gradients across all time scales. However, beyond $d{>}300$, we observe exponential decay of dependence, where the value of MI falls below $10^{-5}$ indicating no further dependencies. This point ($d{\approx}780$) indicates the span of the dependencies.

\subsubsection{Experiments with Dilated Recurrent Neural Networks}\label{ssec:drnn}
Recurrent neural architectures always deliver better performance on unpermuted sequential MNIST as compared to permuted sequential MNIST. Our analysis of the dependency decay curves of these datasets provides an explanation for why this is the case. Unpermuted sequential MNIST has LDDs of less than $30$, due to the period nature of pixel dependencies as explained in section~\ref{ssec:ldds_sequential_mnist}. Datasets possessing such short-range dependency can be easily modeled using simple models as they don't require long memory. In the case of permuted sequential MNIST, we observe LDDs of more than $780$ (due to exponential decay beyond that). Here, we use DilatedRNNs to train unpermuted and permuted sequential MNIST datasets in a classification task (classify digits $0$-$9$ from their images). DilatedRNNs use multi-resolution dilated recurrent skip connections to extend the range of temporal dependencies in every layer and upon stacking multiple such layers are able to learn temporal dependencies at different scales \cite{chang2017dilated}. This stacking of multi-resolution layers helps in passing contextual information over long distances which otherwise would have vanished via a single layer. Thus, the set of the dilations should, ideally, be tailored to match the dependency decay curves present in the dataset, and, in particular, the max dilation should match the max LDDs present in the dataset. The dilations per layer, and the  number of layers, within a DilatedRNN are controlled by hyper-parameters \cite{chang2017dilated}. 

\begin{table}
\small
\centering
\caption{Results for sequential MNIST using GRU cells}
\label{tab:gru_unper}
\begin{tabular}{cccc}
\hline
\# of & Set of & Hidden & Accuracy \\
 Layers & Dilations & per Layer & \\
\hline
$4$ & $1,2,4,8$ & $20/50$ & $98.96/99.18$ \\
$5$ & $1,2,4,8,16$ & $20/50$ & $98.94/99.21$ \\
$6$ & $1,2,4,8,16,32$ & $20/50$ & $\mathbf{99.17/99.27}$ \\
$7$ & $1,2,4,8,16,32,64$ & $20/50$ & $99.05/99.25$ \\
$8$ & $1,2,4,8,16,32,64,128$ & $20/50$ & $99.15/99.23$ \\
$9$ & $1,2,4,8,16,32,64,128,256$ & $20/50$ & $98.96/99.17$ \\
\hline
\end{tabular}
\end{table}

\begin{table}
\small
\centering
\caption{Results for permuted sequential MNIST using RNN cells}
\label{tab:rnn_per}
\begin{tabular}{cccc}
\hline
\# of & Set of & Hidden & Accuracy \\
 Layers & Dilations & per Layer & \\
\hline
$7$ & $1,2,4,8,16,32,64$ & $20/50$ & $95.04/95.94$ \\
$8$ & $1,2,4,8,16,32,64,128$ & $20/50$ & $95.45/95.88$ \\
$9$ & $1,2,4,8,16,32,64,128,256$ & $20/50$ & $95.5/96.16$ \\
$10$ & $1,2,4,8,16,32,64,128,256,512$ & $20/50$ & $95.62/96.4$ \\
$11$ & $1,2,4,8,16,32,64,128,256,512,780$ & $20/50$ & $\mathbf{95.66/96.47}$ \\
\hline
\end{tabular}
\end{table}

The original paper that introduced DilatedRNNs~\cite{chang2017dilated} used the same max dilation hyper-parameter for both of these datasets i.e. $256$, and a standard set of dilations (i.e., $1,2,4,8,\dots$). The best results reported by~\cite{chang2017dilated} for these two datasets were: unpermuted sequential MNIST $99.0$/$99.2$ and permuted sequential MNIST $95.5$/$96.1$. However, our analysis of these datasets has revealed different max dependencies across these dataset. For unpermuted sequential MNIST we identified a periodicity of $28$ and so we expected the max dilation value to be near $28$ to deliver better performance. In permuted sequential MNIST we identified that the dependencies extend up to $780$ and so we would expect better performance by extending the max dilation up to this value. To test these hypotheses we trained DilatedRNNs with various sets of dilations. To keep our results comparable with those reported in~\cite{chang2017dilated} the original code\footnote{\url{https://github.com/code-terminator/DilatedRNN}} was kept unchanged except for the max dilation hyper-parameter. The test results of these experiments are in tables~\ref{tab:gru_unper} and~\ref{tab:rnn_per}. For unpermuted task, the model delivered best performance for max dilation of $32$. Focusing on the results of the permuted sequential MNIST, the best performance was delivered with the max dilation of $780$. These results confirm that the best performance is obtained when the max dilation is similar to the span of the LDDs of a given dataset.

\section{Computing Dependency Decay Curves of Artificial Datasets}
Natural datasets present little to no control over the factors the affect LDDs. This, limits our ability to understand LDDs in more detail. Strictly \emph{k}-Piecewise Languages (SP\emph{k}) languages exhibit some types of LDDs occurring in natural datasets. Moreover, by modifying the SP\emph{k} grammar we can control the dependency decay curves within a dataset generated by the grammar. To understand and validate the interaction between an SP\emph{k} grammar and the LDD properties of the data we use a number of SP\emph{k} grammars to generate datasets and analyse the properties of these datasets. Below is the description of the SP\emph{k} language.

\subsection{Strictly \emph{k}-Piecewise Languages (SP\emph{k})}\label{sec:spk_def}
SP\emph{k} languages form a subclass of regular languages. Subregular languages can be identified by mechanisms much less complicated than Finite-State Automata. Many aspects of human language such as local and non-local dependencies are similar to subregular languages \cite{Jager2012}. More importantly, there are certain types of long distance (non-local) dependencies in human language which allow finite-state characterization \cite{Heinz2010}. These type of LDDs can easily be characterized by SP\emph{k} languages and can be easily extended to other processes.

A language \emph{L}, is described by a finite set of unique symbols \( \Sigma \) and \( \Sigma \)* (\emph{free monoid}) is a set of finite sequences or strings of zero or more elements from \(\Sigma \).

\theoremstyle{definition}
\begin{exmp}
Consider, \(\Sigma \) = \{\emph{\( \sigma \)\textsubscript{1}, \( \sigma \)\textsubscript{2}, \( \sigma \)\textsubscript{3}, \( \sigma \)\textsubscript{4}}\} where \emph{\( \sigma \)\textsubscript{1}, \( \sigma \)\textsubscript{2}, \( \sigma \)\textsubscript{3}, \( \sigma \)\textsubscript{4}} are the unique symbols. A \emph{free monoid} over \(\Sigma \) contains all concatenations of these unique symbols. Thus, \( \Sigma \)* = \{\emph{\( \lambda \), \( \sigma \)\textsubscript{1}, \( \sigma \)\textsubscript{1}\( \sigma \)\textsubscript{2}, \( \sigma \)\textsubscript{1}\( \sigma \)\textsubscript{3}, \( \sigma \)\textsubscript{1}\( \sigma \)\textsubscript{4}, \( \sigma \)\textsubscript{3}\( \sigma \)\textsubscript{2}, \( \sigma \)\textsubscript{3}\( \sigma \)\textsubscript{1}\( \sigma \)\textsubscript{3}, \( \sigma \)\textsubscript{2}\( \sigma \)\textsubscript{1}\( \sigma \)\textsubscript{4}\( \sigma \)\textsubscript{3}, ... }\}.
\end{exmp}

\theoremstyle{definition}
\begin{definition}
Let, \emph{u} denote a string, e.g. \emph{u}= \( \sigma \)\textsubscript{3}\( \sigma \)\textsubscript{2}. The length of a string \emph{u} is denoted by \( \vert u \vert \), and if \emph{u}= \( \sigma \)\textsubscript{3}\( \sigma \)\textsubscript{2} then \( \vert u \vert \)=2. A string with length zero is denoted by \( \lambda \).
\end{definition}

\theoremstyle{definition}
\begin{definition}\label{def1}
A string \emph{v} is a \emph{subsequence} of string \emph{w}, iff \emph{v} = \( \sigma \)\textsubscript{1}\( \sigma \)\textsubscript{2} ... \( \sigma \)\textsubscript{n} and \emph{w} \( \in \Sigma \)*\( \sigma \)\textsubscript{1}\( \Sigma \)*\( \sigma \)\textsubscript{2}\( \Sigma \)* ... \( \Sigma \)*\( \sigma \)\textsubscript{n}\( \Sigma \)*, where \( \sigma \in \Sigma\). A \emph{subsequence} of length \emph{k} is called a \emph{k-subsequence}. Let subseq\textsubscript{\emph{k}}(\emph{w}) denote the set of subsequences of \emph{w} up to length \emph{k}.
\end{definition}

\theoremstyle{definition}
\begin{exmp}
Consider, \( \Sigma \) = \{\emph{a, b, c, d}\}, \emph{w} = [\emph{acbd}], \emph{u} = [\emph{bd}], \emph{v} = [\emph{acd}] and \emph{x} = [\emph{db}]. String \emph{u} is a \emph{subsequence} of length \emph{k} = 2 or \emph{2-subsequence} of \emph{w}. String \emph{v} is a \emph{3-subsequence} of \emph{w}. However, string \emph{x} is \emph{not a subsequence} of \emph{w} as it does not contain [\emph{db}] subsequence.
\end{exmp}

SP\emph{k} languages are defined by grammar \emph{G}\textsubscript{\emph{SPk}} as a set of permissible \emph{k}-\emph{subsequences}. Here, \emph{k} indicates the number of elements in a dependency. Datasets generated to simulate 2 elements in a dependency will be generated using SP\emph{2}. This is the simplest dependency structure. There are more complex chained-dependency structures which require higher \emph{k} grammars.

\theoremstyle{definition}
\begin{exmp}\label{exp1}
Consider \emph{L}, where \( \Sigma \) = \{\emph{a, b, c, d}\}. Let \emph{G\textsubscript{SP\emph{2}}} be SP\emph{k} grammar which is comprised of permissible \emph{2-subsequences}. Thus, \emph{G\textsubscript{SP\emph{2}}} = \{\emph{aa, ac, ad, ba, bb, bc, bd, ca, cb, cc, cd, da, db, dc, dd}\}. \emph{G\textsubscript{SP\emph{2}}} grammar is employed to generate SP\emph{2} datasets.
\end{exmp}

\theoremstyle{definition}
\begin{definition}
Subsequences which are not in the grammar \emph{G} are called \emph{forbidden subsequences}\footnote{Refer section \emph{5.2. Finding the shortest forbidden subsequences} in \cite{Fu2011} for method to compute \emph{forbidden sequences} for SP\emph{k} language}.
\end{definition}

\theoremstyle{definition}
\begin{exmp}
Consider Example~\ref{exp1}, although \{\emph{ab}\} is a possible \emph{2-subsequence}, it is not part of the grammar \emph{G\textsubscript{SP\emph{2}}}. Hence, \{\emph{ab}\} is a \emph{forbidden subsequence}.
\end{exmp}

\theoremstyle{definition}
\begin{exmp}\label{exp2}
Consider strings \emph{u, v, w}: \emph{u} = [\emph{bbcbdd}], \emph{v} = [\emph{bbdbbbcbddaa}] and \emph{w} = [\emph{bbabbbcbdd}], where \( \vert \)\emph{u}\( \vert \) = 6, \( \vert \)\emph{v}\( \vert \) = 12 and \( \vert \)\emph{w}\( \vert \) = 10. Strings \emph{u} and \emph{v} are valid SP\emph{2} strings because they are composed of subsequences that are in \emph{G\textsubscript{SP\emph{2}}}. However, \emph{w} is an invalid SP\emph{2} string because \emph{w} contains \{\emph{ab}\} a subsequence which is a \emph{forbidden subsequence}. These constraints apply for any string \emph{x} where \( \vert \)\emph{x}\( \vert \in \mathbb{Z} \).
\end{exmp}

\theoremstyle{definition}
\begin{exmp}\label{LDDLength}
Let \emph{G\textsubscript{SP\emph{3}}}  = \{\emph{aaa, aab, abb, baa, bab, bba, bbb, ...}\} and \emph{forbidden subsequence} = \{\emph{aba}\} be an SP\emph{3} grammar which is comprised of permissible $3$-\emph{subsequences}. Thus, \emph{u} = [\emph{aaaaaaab}], where \( \vert \)\emph{u}\( \vert \) = 8 is a valid SP\emph{3} string and \emph{v} = [\emph{aaaaabaab}], where \( \vert \)\emph{v}\( \vert \) = 9 is an invalid SP\emph{3} string as defined by the grammar \emph{G\textsubscript{SP\emph{3}}}.
\end{exmp}

The maximum extent of LDD exhibited by a certain SP\emph{k} language is equal to the length of the strings generated which abide by the grammar. However, as per definition~\ref{def1}, the strings generated using this method will also exhibit dependencies of shorter lengths. It should be noted that the length of the LDD is not the same as \emph{k}. The length of the LDD is the maximum distance between two elements in a dependency, whereas \emph{k} specifies the number of elements in the dependency (as defined in the the SP\emph{k} grammar).

\begin{exmp}
As per Example~\ref{exp2}, \emph{v} = [\emph{bbdbbbcbddaa}], consider \emph{b} in the first position, \emph{subsequence} \{\emph{ba}\} exhibits dependency of 10 and 11. Similarly, \emph{subsequence} \{\emph{bd}\} exhibits dependency of 2, 9, 10, etc.
\end{exmp}

\begin{figure*}
\centering
\includegraphics[scale=0.4]{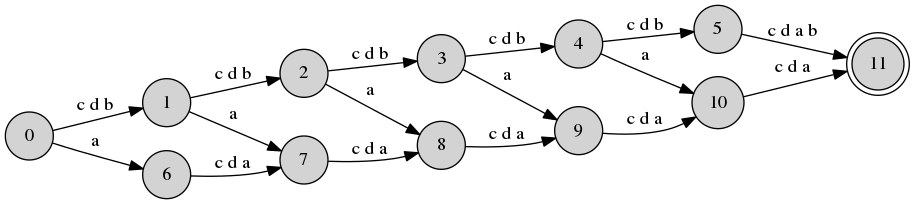}
\caption{The automaton for \emph{G\textsubscript{SP2}} where \emph{n\textsubscript{l}=6}}
\label{fig:fsa1}
\end{figure*}

Fig.~\ref{fig:fsa1} depicts a finite-state diagram of \emph{G\textsubscript{SP2}}, which generates strings of synthetic data. Consider a string \emph{x} from this data, \( \forall \) generated strings \emph{x} generated using grammar \emph{G\textsubscript{SP2}}: \( \vert \)\emph{x}\( \vert \) = $6$. The \emph{forbidden subsequence} for this grammar is \{\emph{ab}\}. Since \{\emph{ab}\} is a \emph{forbidden subsequence}, the state diagram has no path (from state $0$ to state $11$) because such a path would permit the generation of strings with \{\emph{ab}\} as a subsequence, \emph{e.g.} \{\emph{abcccc}\} 
Traversing the state diagram generates valid strings \emph{e.g.} \{\emph{accdda, caaaaa}\}.

Various \emph{G\textsubscript{SP\emph{k}}} could be used to define an SP\emph{k} depending on the set of \emph{forbidden subsequences} chosen. Thus, we can construct rich datasets with different properties for any SP\emph{k} language. \emph{forbidden subsequences} allow for the elimination of certain logically possible sequences while simulating a real world dataset where the probability of occurrence of that particular sequence is highly unlikely. Every SP\emph{k} grammar is defined with at least one \emph{forbidden subsequence}.

\subsection{Dependency decay curves of SP\emph{k} datasets}
\begin{figure}
\centering
\subcaptionbox{\label{fig:spk_len}}{
   \includegraphics[scale=0.43]{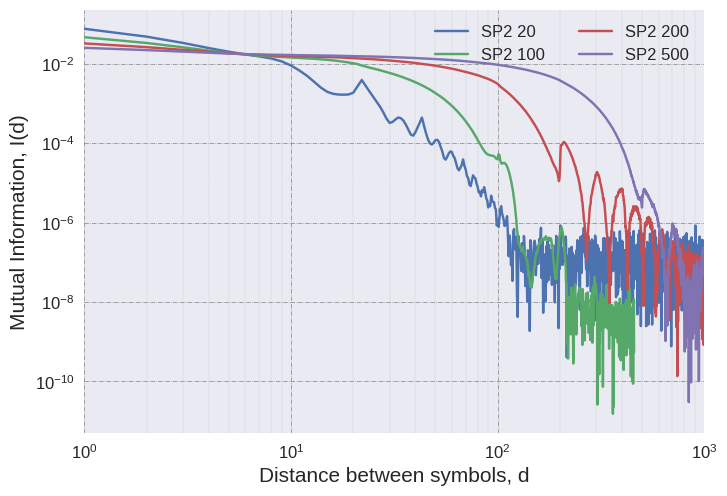}
}\hfill
\subcaptionbox{\label{fig:spk_k}}{
	\includegraphics[scale=0.43]{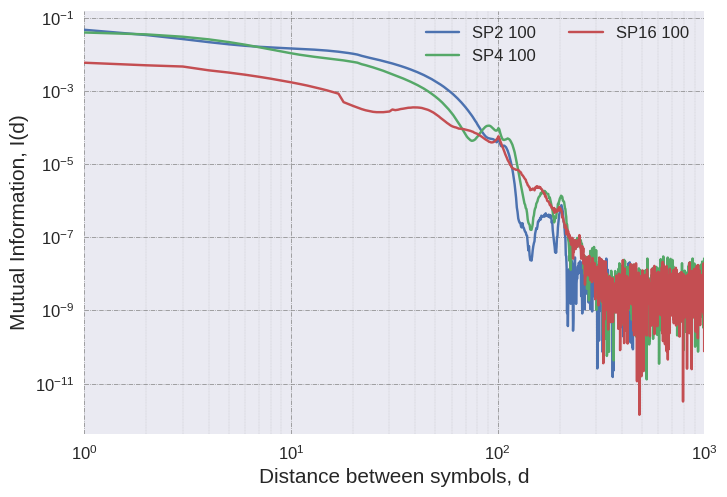}
}\hfill
\subcaptionbox{\label{fig:spk_v}}{
    \includegraphics[scale=0.43]{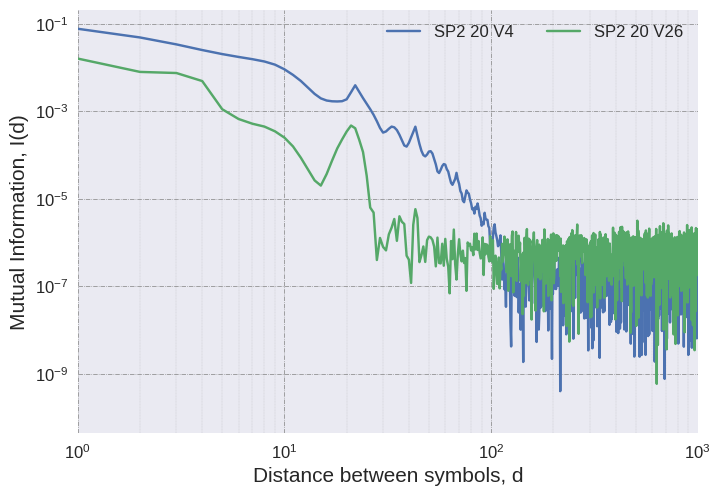}
}\hfill
\subcaptionbox{\label{fig:spk_f}}{
  \includegraphics[scale=0.43]{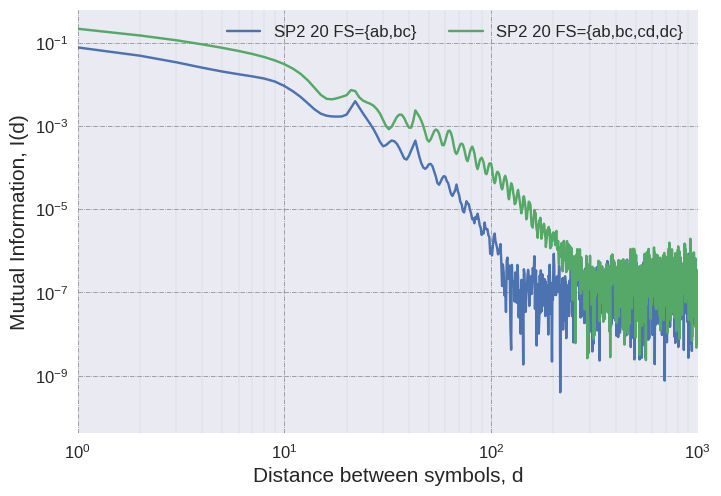}
}\hfill
\subcaptionbox{\label{fig:spk_size}}{
    \includegraphics[scale=0.43]{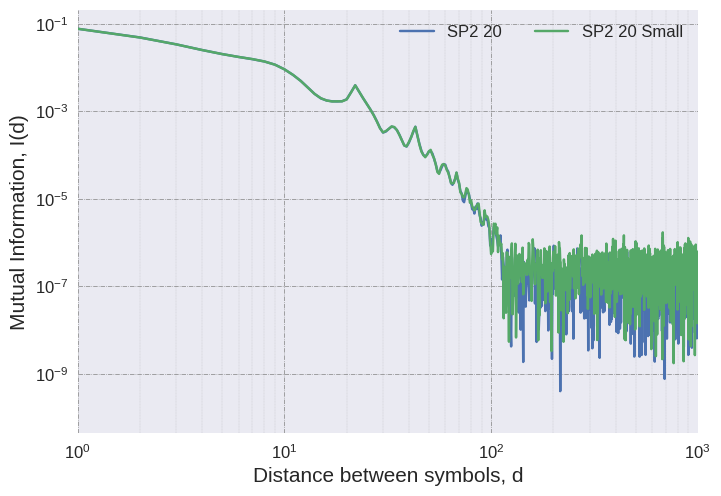}
}\hfill
\subcaptionbox{\label{fig:spk_full}}{
	\includegraphics[scale=0.43]{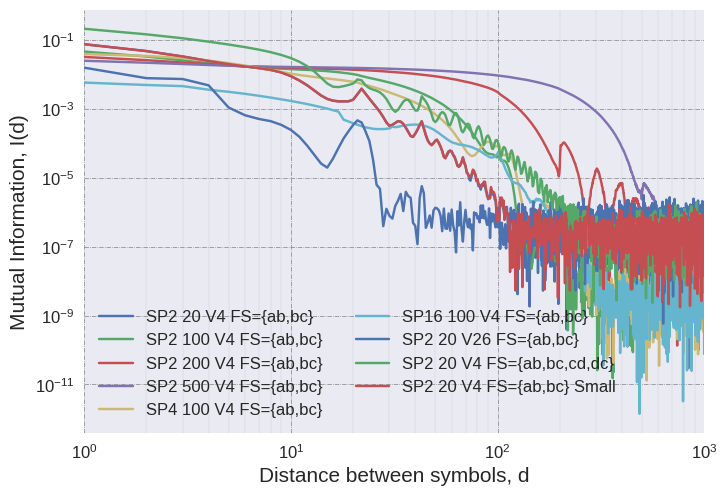}
}\hfill
\caption{Dependency decay curves of datasets of (a) SP\emph{2} grammar exhibiting LDDs of length $20$, $100$, $200$, and $500$. (b) SP\emph{2}, SP\emph{4} and SP\emph{16} grammar exhibiting LDD of length $100$. (c) SP\emph{2} grammar with vocabulary of $4$ and $26$. (d) SP\emph{2} grammar with varying forbidden strings (e) SP\emph{2} grammar with varying size of the datasets (f) All the SP\emph{k} grammars in a plot}
\label{fig:spk}
\end{figure}

In-order to analyse the impact of vocabulary size on dependency decay curves, we generated SP\emph{2} grammars where $\Sigma_1 = \{ a,b,c,d \}$ (size of vocabulary = $4$) and $\Sigma_2 = \{ a,b,c,d,....,x,y,z \}$ (size of vocabulary = $26$). We generated strings of maximum length of $20, 100, 200$ and $500$ using SP\emph{2} grammar. As explained in Example~\ref{LDDLength}, by increasing the length of the generated strings, the distance between dependent elements is also increased, resulting in longer LDDs. We can then simulate LDD lengths as $20, 100, 200$ and $500$. We also choose two sets of forbidden strings for SP\emph{2} grammar, \{$ab, bc$\} and \{$ab, bc, cd, dc$\}. We also generate two sizes of the same SP\emph{2} grammar to study the impact of the size of the data on the dependency decay curves, where one dataset was twice the size of the other. The datasets were generated using \emph{foma} \cite{hulden2009} and \emph{python} \cite{Mahalunkar2018}. Fig.~\ref{fig:spk} shows plots of the dependency decay curves of these datasets.

\subsubsection{Impact of Dependency Length}
Fig.~\ref{fig:spk_len} plots dependency decay curves of SP\emph{2} languages with maximum string length of {$20, 100, 200$, and $500$}. The point where dependency decay is faster, the \emph{inflection point}, lies around the same point on \emph{x-axis} as the maximum length of the LDD. This confirms that SP\emph{k} can generate datasets with varying lengths of LDDs. 

\subsubsection{Impact of \emph{k} - Multi-Element Dependency}
Fig.~\ref{fig:spk_k} plots the dependency decay curves of SP\emph{2}, SP\emph{4} and SP\emph{16} grammars. The strings in all the grammars are up to $100$. Hence, we can observe the mutual information decay beyond $D {>} 100$. \emph{k} defines the number of correlated or dependent elements in a dependency rule. As \emph{k} increases the grammar becomes more complex and there is an overall reduction in frequency of the dependent elements in a given sequence (due to lower probability of these elements occurring in a given sequence). Hence, the mutual information is lower. This can be seen with dataset of SP\emph{16} vs SP\emph{2} and SP\emph{4}. It is worth noting that datasets with lower mutual information curves tend to present more difficulty during modeling \cite{Mahalunkar2018}.

\subsubsection{Impact of Vocabulary Size}
The impact of vocabulary size can be seen in Fig.~\ref{fig:spk_v} where the dependency decay curves of SP\emph{2} datasets with vocabulary size ($V$) $4$ and $26$ are plotted. Both these datasets contain strings of maximum length 20. Hence the mutual information decays at 20. Both curves have identical decay indicating a similar grammar. However the overall mutual information of the dataset with $V=26$ is much lower then the mutual information of the dataset with $V=4$. This is because a smaller vocabulary results in an increase in the probability of the occurrence of each elements of occurrence of elements.

\subsubsection{Impact of Forbidden Strings}
Fig.~\ref{fig:spk_f} plots the dependency decay curves of SP\emph{2} grammar with two set of \emph{forbidden strings} as \{$ab, bc$\} and \{$ab, bc, cd, dc$\}. It is seen that the dataset with more forbidden strings exhibited less steeper mutual information decay than the one with less number of forbidden strings. This can be attributed to the fact that datasets with more complex forbidden strings tend to exhibit more complex grammar as explained in section~\ref{sec:spk_def}. By introducing more number of forbidden strings, it is possible to synthesize more complex LDDs as seen in the plot. In Fig.~\ref{fig:spk_size} we can observe the impact of the size of the dataset sampled from the same grammar. It can be seen that datasets sampled from the same grammar are less likely to be affected by the size of the dataset.

\section{Discussion}
The dependency decay curves of a dataset enable the visualisation of certain type of grammar in the dataset. For example, our analysis of word-based and character-based datasets in sections~\eqref{ssec:ldds_word-based} and~\ref{ssec:ldds_char-based}, indicate that the word-based grammar is very different from character-based grammar. Understanding the properties of the underlying grammar that produces a sequence can aid in choosing best recurrent neural architecture to learn that grammar. For example, the maximum length of LDDs is much smaller in word-based datasets as compared to character-based datasets. But at the same time word-based dependency decay curves exhibit higher value of overall mutual information. This is why the sequential model that performs best on the word-based language modeling task will not necessarily be the best choice for the character-based language modeling task. 

One implication of these experiments is that having multiple benchmark datasets from a single domain does not necessarily improve the experimental testing of a models ability to learn LDDs in the dataset. LDDs are fixed within a domain and sampling more datasets from that domain simply results in testing the model on LDDs with similar characteristics. Consequently, the relatively limited set of domains and tasks covered by benchmark datasets indicates that current benchmarks do not provide enough LDD variety to extensively test the capacity of recurrent neural architectures.

It can also be noted that even though a specific grammar does induce similar dependency decay curves, there are subtle variations. These variations depend on a number of factors such as size of the vocabulary, size of the dataset, dependency structure (for e.g. ``\emph{k}" and ``\emph{forbidden strings}") and presence of any other noisy data (or presence of another grammar as seen with \emph{Enwik8} dataset). Thus, if a recurrent neural architecture intends to model a dataset, knowing these factors would greatly benefit in selecting the best hyper-parameters of the sequential model. Also, these artificial grammars allow for the generation of rich datasets by setting the parameter \emph{k}, the maximum length of the strings generated, size of vocabulary and by choosing appropriate \emph{forbidden substrings}. This presents a compelling case to use these grammars to benchmark state-of-the-art sequential models.

As seen in dependency decay curves of sequential MNIST, it is evident that the use of sequential MNIST in benchmark tasks is out of place due to the absence of long distance dependencies. This presents a compelling case to analyze dependency decay curves of benchmark datasets before they are selected for this job. Also, the presence of a flat curve with very low mutual information in permuted sequential MNIST dataset is usually a result of noisy data. It is this noisy data (rather than complex LDDs) that is responsible for reducing the performance of the recurrent neural architectures. Overall, however we would argue that both of these datasets are inadequate at benchmark sequential models due to their limitation in generating complex LDDs.

\section{Conclusion}
The major contribution of this paper represent a synthesis of distinct themes of research on LDDs from multiple fields, including information theory, artificial neural networks for sequential data modeling, and formal language theory. The potential impact of this synthesis for neural networks research include: an appreciation of the multifaceted nature of LDDs; a procedure for measuring dependency decay curves within a dataset; an evaluation and critique of current benchmark datasets and tasks for LDDs; an analysis of how the use of these standard benchmarks and tasks can be misleading in terms of evaluating the capacity of a neural architectures to generalize to datasets with different forms of LDDs; and, a deeper understanding of the relationship between hyper-parameters and LDDs within language model architectures which can directly contribute to the development of more accurate and efficient sequential models.

\section*{Acknowledgment}
This research was partly supported by the ADAPT Research Centre, funded under the SFI Research Centres Programme (Grant 13/RC/2106) and is co-funded under the European Regional Development Funds. We gratefully acknowledge the support of NVIDIA Corporation with the donation of the Titan Xp GPU under NVIDIA GPU Grant used for this research.

\bibliographystyle{unsrt}  
\bibliography{ldds,anthology}

\end{document}